\documentclass{ieeeaccess}
\usepackage{cite}
\usepackage{academicons}
\usepackage{algorithm}
\usepackage{algpseudocode}
\usepackage{amsfonts}
\usepackage{amsmath, amssymb, amsfonts}
\usepackage{amsthm, enumitem}
\usepackage{array}
\usepackage{booktabs}
\usepackage{caption}
\usepackage{color}
\usepackage{comment}
\usepackage{float}
\usepackage[T1]{fontenc}
\usepackage{graphicx}
\usepackage{hyperref}
\usepackage[utf8]{inputenc}
\usepackage{lipsum}
\usepackage{longtable}
\usepackage{multirow}
\usepackage{stfloats}
\usepackage{subcaption}
\usepackage{mathptmx}
\usepackage{textcomp}
\usepackage[flushleft]{threeparttable}
\usepackage{url}
\usepackage{wrapfig}
\def\orcid#1{\kern .08em\href{https://orcid.org/#1}{\includegraphics[keepaspectratio,width=0.7em]{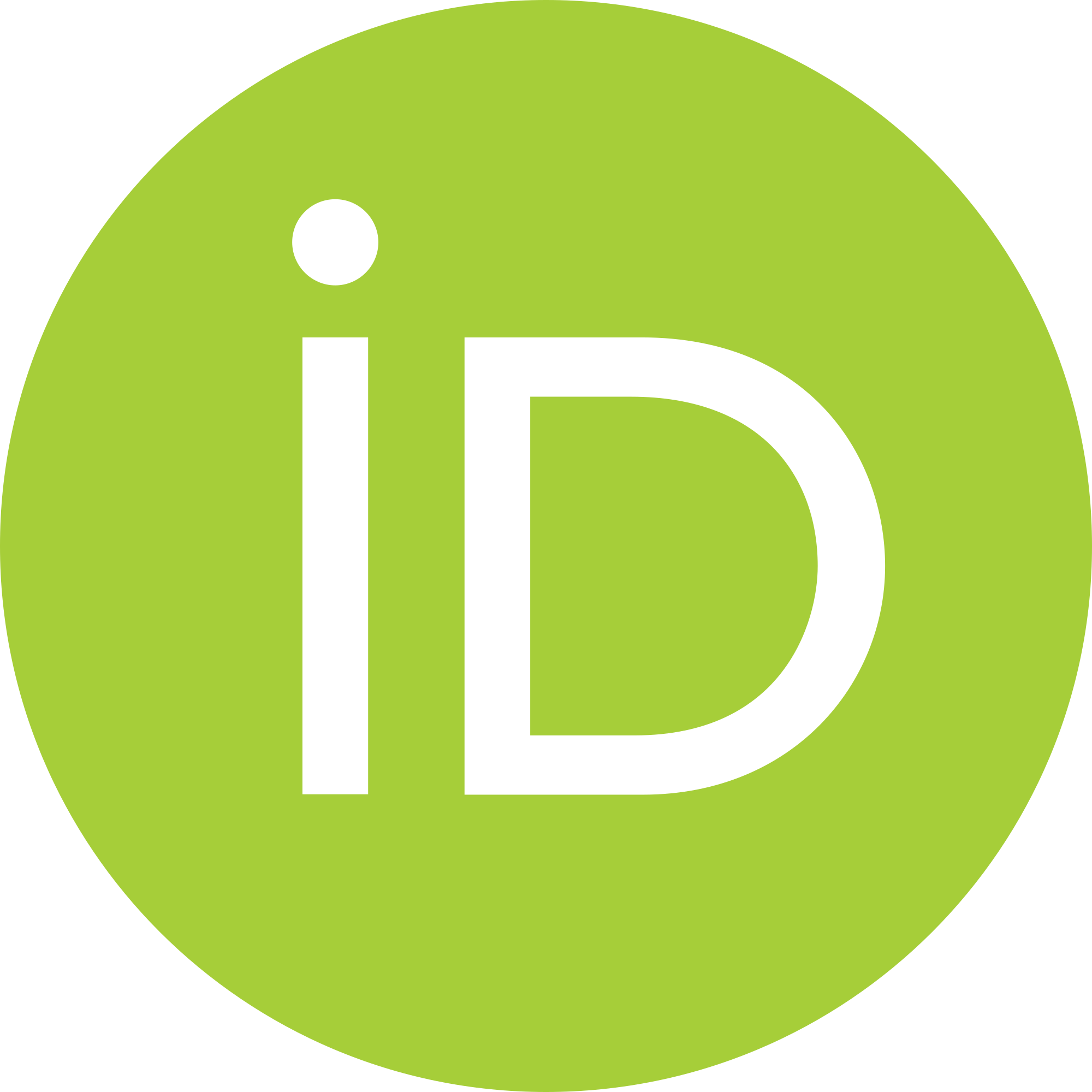}}} 
\usepackage{verbatim}
\usepackage{setspace}
\floatstyle{plaintop}

\begin{document}

\history{Date of publication xxxx 00, 0000, date of current version xxxx 00, 0000.}
\doi{10.1109/ACCESS.2023.0322000}

\title{PCA-Featured Transformer for Jamming Detection in 5G UAV Networks}

\author{
{Joseanne Viana}\authorrefmark{1, 3}\textsuperscript{\textsection} \orcid{0000-0002-4191-3127} \and \IEEEmembership{IEEE Member},
{Hamed Farkhari}\authorrefmark{2}\textsuperscript{\textsection} \orcid{0000-0002-2620-260X} \and \IEEEmembership{IEEE Member},
{Pedro Sebastião} \authorrefmark{2,4}\orcid{0000-0001-7729-4033} \and\IEEEmembership{IEEE Member}, 
{Victor P Gil Jimenez}\authorrefmark{1} \orcid{0000-0001-7029-1710} \and \IEEEmembership{IEEE Senior Member}}

\address[1]{UC3M - Universidad Carlos III de Madrid, Madrid, Spain;} 
\address[2]{ISCTE – Instituto Universitário de Lisboa, Av. das Forças Armadas, 1649-026 Lisbon, Portugal;}
\address[3]{Tyndall National Institute, Ireland; }
\address[4]{IT – Instituto de Telecomunicações, Av. Rovisco Pais, 1, Torre Norte, Piso 10, 1049-001 Lisboa, Portugal;}

\corresp{Corresponding author: Joseanne Viana (joseanne.viana@tyndall.ie)}

\begin{abstract}
Unmanned Aerial Vehicles (UAVs) face significant security risks from jamming attacks, which can compromise network functionality. Traditional detection methods often fall short when confronting AI-powered jamming that dynamically modifies its behavior, while contemporary machine learning approaches frequently demand substantial feature engineering and struggle with temporal patterns in attack signatures. The vulnerability extends to 5G networks employing Time Division Duplex (TDD) or Frequency Division Duplex (FDD), where service quality may deteriorate due to deliberate interference.
We introduce a novel U-shaped transformer architecture that leverages Principal Component Analysis (PCA) to refine feature representations for improved wireless security. The training process is regularized by incorporating the output entropy uncertainty into the loss function, a mechanism inspired by the Soft Actor-Critic (SAC) algorithm in Reinforcement Learning (RL) to enable robust jamming detection techniques. 
The architecture features a modified transformer encoder specially designed to process critical wireless signal features, including Received Signal Strength Indicator (RSSI) and Signal-to-Interference-plus-Noise Ratio (SINR) measurements. We complement this with a custom positional encoding mechanism that specifically accounts for the inherent periodicity of wireless signals, enabling a more accurate representation of temporal signal patterns.
In addition, we propose a batch size scheduler and implement chunking techniques to optimize training convergence for time series data. These advancements contribute to achieving up to a ten times improvement in training speed within the advanced U-shaped encoder-decoder transformer model introduced in this study.
Experimental evaluations demonstrate the effectiveness of our entropy-based approach, achieving detection rates of 89.46\% under Line-of-Sight (LoS) conditions and 85.06\% in non-Line-of-Sight (NLoS) scenarios. Our method significantly outperforms existing solutions, surpassing XGBoost (XGB) classifiers by approximately 4.5\% and other deep learning approach by more than 2\%.

\end{abstract}

\setlength{\textfloatsep}{4pt }
\maketitle
\begingroup\renewcommand\thefootnote{\textsection}
\footnotetext{Collaborative authors with equal contribution}
\endgroup

\IEEEoverridecommandlockouts

\IEEEpeerreviewmaketitle

\section{Introduction}

The intersection of Unmanned Aerial Vehicles (UAVs) and wireless communication systems represents a rapidly evolving research domain with significant technological and security implications. As UAVs transition from specialized military applications to widespread commercial deployment, their integration into communication networks introduces novel challenges and opportunities \cite{Bekkouche2020}, \cite{Naqvi2018}, \cite{Azari2019}, \cite{Wenbo2020}, \cite{Geraci2022}, \cite{Qi2019}. Base stations mounted on UAVs demonstrate potential for applications including emergency response, surveillance of borders, and providing temporary network coverage \cite{Ho24}, \cite{Galkin2023}, \cite{Kaleem2018}, \cite{Wenbo2020}. However, when UAVs function as end-devices in services such as package delivery, they introduce distinct security vulnerabilities within 5G networks \cite{Xiao2018}, \cite{Lee2018}. A critical concern is the vulnerability of UAV communication systems to sophisticated jamming attacks, which can manipulate Time Division Duplex (TDD) and Frequency Division Duplex (FDD) systems, resulting in severe service disruptions with impacts reaching up to 99\% in TDD uplink and 82\% in FDD downlink scenarios \cite{Skokowski2024}, \cite{Viana2024}, \cite{Darsena2022}. Jamming can affect private networks, compromising the integrity and availability of mission critical communications in industrial, corporate, and specialized operational environments where UAVs are increasingly deployed.

Machine learning, particularly deep learning approaches, offers promising avenues for developing more effective and proactive jamming detection systems \cite{Xin2018}, \cite{Berman2019}, \cite{Mao2018}. Convolutional Neural Networks (CNNs) have shown effectiveness in extracting spatial features from signal data \cite{Naseer2018}, \cite{Krizhevsky2012}, \cite{Fran2016}, while Long Short-Term Memory (LSTM) networks excel at modeling temporal patterns in wireless communications \cite{Zhao2017}, \cite{Ismail2019}, \cite{VianaConv2022}, \cite{Farkhari2024}.

Several studies have advanced this field. \cite{Ju2019} developed a branched deep neural network architecture for simultaneous jamming detection and link scheduling in dense wireless networks. Their system employs two specialized subnetworks: one leverages geographical information and signal power measurements to detect and locate jammers, while the other optimizes link scheduling based on jamming detection results to maximize network throughput under adverse conditions. \cite{RuoRan2016} proposed neural networks for compound jamming signal recognition, while \cite{Li2021} introduced a singular value decomposition approach for jamming identification in Global Navigation Satellite System (GNSS)-based systems.

For radar systems specifically, \cite{Lv2022} introduced a deep learning method to counter interrupted sampling deceptive jamming. Their four-stage approach begins by generating time-frequency representations of radar echoes using short-time Fourier transform, which feed into a You Only Look Once (YOLO) object detection model that identifies jammed signals while implementing position correction to preserve legitimate target information. \cite{Gao2021} further expanded this domain by proposing deep neural networks for digital radio frequency memory (DRFM) jamming mode identification.

Addressing cognitive radio vulnerabilities, \cite{Aygul2020} created a one-dimensional convolutional neural network operating directly on raw signal data to detect primary user emulation and jamming attacks. This approach eliminates the manual feature engineering required by traditional methods, as noted by \cite{Arjoune2020}. Their architecture incorporates three convolutional layers with rectified linear unit (ReLU) activation functions, followed by dense and softmax output layers.

More recently, \cite{Amini2025} developed a deep learning system using ensemble techniques for detecting jamming attacks in 5G networks, combining RF domain and physical layer features with a Temporal Epistemic Decision Aggregator to enhance detection reliability despite signal impairments and carrier frequency offset. Similarly, \cite{Li2022} proposed a feature- and spectrogram-tailored machine learning approach for jamming detection in OFDM-based UAVs.

Despite these advances, these architectures often struggle to capture the complex interdependencies and multifaceted patterns characteristic of modern jamming attacks, particularly in heterogeneous environments that incorporate both 5G New Radio (NR) and Narrowband Internet of Things (NB-IoT) interfaces \cite{Hachimi2020}, \cite{Krayani2022}, \cite{Viana2024}, \cite{Sahu2021}.

Transformer architectures, with their self-attention mechanisms \cite{Vaswani2019}, have improved sequential data analysis and offer significant potential for jamming detection \cite{KaiJui2022}. These models are particularly well-suited for identifying long-range dependencies in wireless signal data, enabling the detection of sophisticated jamming patterns across various time scales and frequency bands \cite{10654316}. The multi-head attention mechanism further enhances this capability by simultaneously analyzing diverse signal attributes, from immediate interference to subtle, persistent disruptions, making transformers especially effective against energy-efficient selective jamming techniques \cite{Elleuch2024}.

With the accelerating deployment of 5G networks and increasingly complex security threats \cite{liyanage2018comprehensive}, \cite{Schneider2015}, there is a need for advanced jamming detection methods that can preemptively identify and counteract these attacks \cite{Xue2023}, \cite{Lin2022}. Integrating transformers with other machine learning techniques offers promising new approaches for addressing these challenges, establishing foundations for more resilient network security frameworks \cite{Mowla2020}, \cite{Sun2018}.

This paper introduces an innovative transformer-based architecture specifically designed for UAV-integrated 5G networks, focusing on early detection of jamming attacks \cite{VianaConv2022}, \cite{Viana2024}. By incorporating Principal Component Analysis (PCA)-derived features \cite{Greenacre2022}, \cite{Farkhari2022}, our approach enables efficient analysis of critical signal metrics, including Received Signal Strength Indicator (RSSI) and Signal to Interference plus Noise Ratio (SINR) \cite{Li2019}, \cite{Geraci2022}. Multi-head attention mechanisms allow the model to identify and classify complex jamming patterns, while computational optimizations make the framework suitable for edge device deployment \cite{Gao2020}, \cite{JVianaAcu2022}. This work not only advances the current state of jamming detection but also proposes techniques to improve training time in machine learning integrated wireless communications \cite{Liu2020}, \cite{Su2020}.

Furthermore, our proposed solution addresses gaps in existing research by prioritizing real-time detection capabilities \cite{Nguyen2008}, \cite{Sperotto2010} through a transformers models. The architecture's adaptability allows it to function effectively across various deployment scenarios, ensuring broad applicability for diverse UAV applications in both civilian and defense contexts \cite{SKOKOWSKI2022}, \cite{Bastos2021}, \cite{Harvey2019}. As UAV technology continues to expand \cite{Kang2022}, \cite{Wang2022}, ensuring communication system security and reliability remains essential. Our work makes a meaningful contribution toward this goal, providing a robust and scalable solution to one of the most pressing challenges in modern wireless communication.

\subsection{\textbf{Contributions and Motivation}}

Recent advancements in cellular networks, particularly in UAV and 5G technologies, have revealed significant vulnerabilities to jamming attacks. While research exists on various detection methods, there remains a critical gap in leveraging transformer architectures for jamming detection. This paper presents the listed key contributions:
\begin{itemize}
\item Developed an innovative transformer-based architecture for detecting jamming attacks in UAV-integrated 5G networks. The system features a custom deep neural network that combines state-of-the-art CNNs with specialized activation functions in a U-Net architecture, optimized for analyzing jamming signatures across 5G NR interfaces.

\item Introduced time-series PCA-features and efficient tokenization method for detecting jamming patterns in UAV-integrated 5G networks.

\item Proposed incorporating the output entropy uncertainty into the loss function.

\item Optimized deep network training algorithm by introducing batch\_size scheduler and chunking (grouping) in the training dataset. 

\item Comprehensive experimental validation demonstrating superior detection capabilities compared to existing methods for jamming attacks.

\end{itemize}

\section{System Model}

The system model comprises an authenticated UAV operating within a small cell network environment designed to detect malicious jamming activities through power variation analysis. In other to train our model, we generated a dataset that has two distinct communication scenarios: Line-of-Sight (LoS) and Non-Line-of-Sight (NLoS). Each scenario category contains four unique experimental configurations that vary key parameters including UAV mobility patterns, operational speeds, attack intensities, and network user density. The dataset architecture specifically accounts for urban environment dynamics, where building structures and other obstacles can significantly impact signal propagation. To ensure robust detection capabilities, we simulate various attack patterns using different numbers of hostile UAVs (ranging from 1 to 4 attackers) with varying transmission powers. The dataset inherently presents an unbalanced distribution between attack and non-attack scenarios, necessitating careful preprocessing to maintain classification accuracy. Each simulation captures temporal sequences of received power measurements of RSSI and SINR. More comprehensive information regarding the dataset, the quantity of jamming UAVs, and signal parameters can be found in \cite{Viana2024} and \cite{SyntheticJViana}.

\subsection{\textbf{Dataset on Jamming Detection}}

\begin{figure}[!ht]
 \centering
 \includegraphics[width=0.50\textwidth]{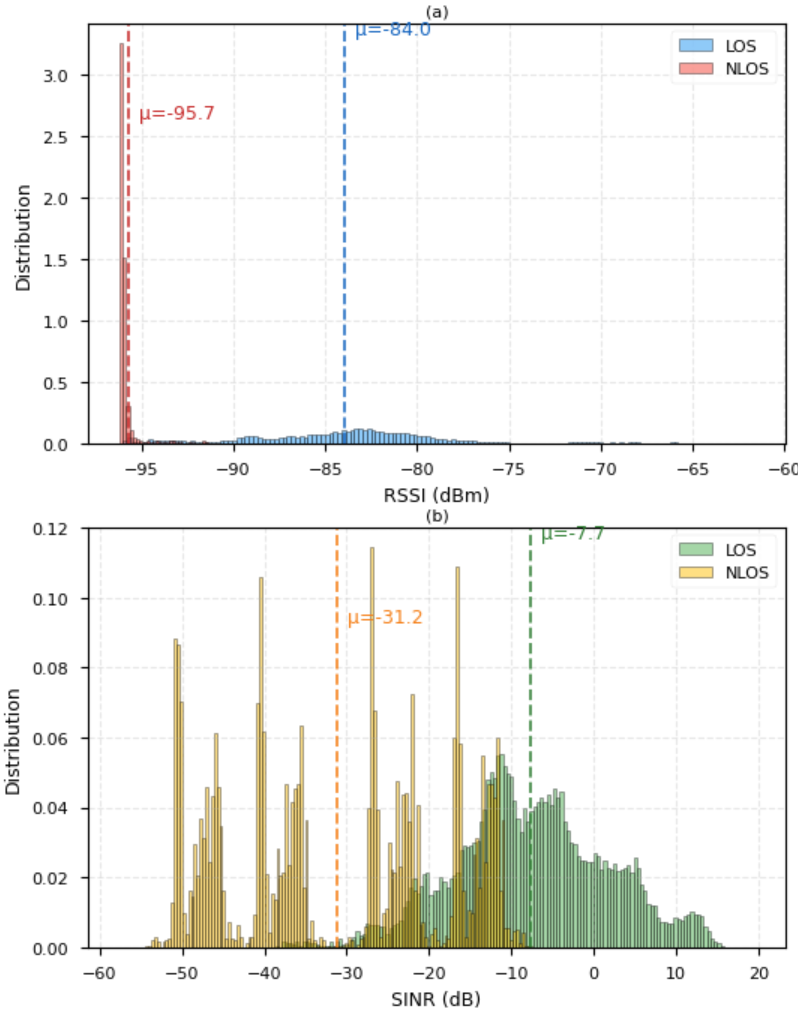}
 \caption{RSSI and SINR with jamming experienced by the UAV receiver.}
 \label{fig:distributions}
\end{figure}

A sample from the dataset on the jamming effects is presented in figure \ref{fig:distributions}, which illustrates the distributions of RSSI and SINR. These simulation results capture the predicted signal propagation characteristics under both LoS and NLoS conditions. 

The RSSI measurements exhibit a distinct bimodal distribution pattern between LoS and NLoS conditions. The NLoS signals demonstrate a concentrated distribution with $\mu=-95.7$ dBm, indicating substantial signal attenuation. Conversely, the LoS signals present a more dispersed distribution centered at $\mu=-84.0$ dBm, reflecting superior signal strength characteristics typical of direct path propagation. The NLoS distribution's pronounced, narrow peak suggests consistent attenuation patterns, while the broader LoS distribution indicates more diverse signal propagation paths despite maintaining direct visibility.

The SINR distributions effectively illustrate the jamming environment's impact on signal quality. NLoS signals exhibit multiple distinct peaks in the range of -50 dB to -20 dB, with a mean value of $\mu=-31.2$ dB, demonstrating severe interference effects. The LoS signals display a more favorable distribution extending into positive values, characterized by $\mu=-7.7$ dB. The broader, right-skewed distribution observed in LoS conditions indicates that despite direct visibility, jamming significantly degrades signal quality, albeit to a lesser extent compared to NLoS scenarios.


\subsection{\textbf{Jamming Detection algorithm for UAVs}}

The proposed approach combines PCA features with transformer architectures to create an efficient and robust detection system.

\subsubsection{\textbf{Feature Engineering with PCA for Time Series Data}}

Direct application of PCA to time series data often fails to capture temporal dependencies effectively, resulting in suboptimal performance. In our initial experiments, training the model solely with PCA components did not yield satisfactory results. To overcome this limitation, we utilized PCA to generate additional features, which were then integrated with the raw data. This approach improved classification accuracy by up to 5\% for both LoS and NLoS datasets.

\paragraph{\textbf{Sample Creation}}
For each time series signal $S$ (either RSSI or SINR), we transformed the 1D signal into a 2D sample matrix. Using a rolling window of size 300, a signal $S$ of length $N$ was converted into a matrix $X$ of shape $(N - 299, 300)$, where each row represents a sample of 300 consecutive time steps.

\paragraph{\textbf{Transformations for Feature Enhancement}}
To capture diverse temporal patterns in the 2D sample matrix $X$, we applied a series of transformations. These include:
\begin{itemize}[noitemsep,topsep=0pt]
 \item The original samples $X$,
 \item Moving averages with window sizes of 2, 3, and 5, adjusted via slicing to align output dimensions,
 \item Sub-sampled versions of $X$ by selecting every 2nd or 3rd feature, with varying starting indices to account for phase shifts.
\end{itemize}
In total, nine distinct transformations per signal were generated. For the precise definitions and implementation details of each transformation, refer to \textbf{Algorithm \ref{alg:feature_creation}} (Feature Creation Algorithm). These transformations enhance the temporal representation of the data, preparing it for the subsequent PCA-based feature extraction step.

\paragraph{\textbf{PCA Application and Feature Extraction}}
For each of the nine transformed matrices:
\begin{enumerate}[noitemsep,topsep=0pt]
 \item A PCA model was fitted to the data, retaining the principal components (PCs) that collectively explain 99\% of the total variance.
 \item For further process, the first five principal components were selected to reducing the feature dimensionality while retaining a significant portion of the data's variability.
\end{enumerate}
This process generated up to five PCA features for each transformation applied to the RSSI and SINR signals. Given that a total of nine transformations were applied to each signal, this resulted in a maximum of 45 features for each original signal (RSSI or SINR). Consequently, for both RSSI and SINR combined, a total of 90 features were obtained for LoS scenarios, whereas for NLoS scenarios, a total of 54 features were derived. This discrepancy arises because, in some transformed signals, only a single PC captured the entire variance of the transformation.

\paragraph{\textbf{Feature Scaling and Integration}}
Each of the 45 PCA feature columns per signal was normalized to the range of the original sample matrix $X$ using the \texttt{MinMaxScaler}, with the feature range set to $(\min(X), \max(X))$. The scaled PCA features were then concatenated with the original samples $X$ along the feature axis, creating an enhanced feature set. This integration preserved the original data structure and avoided modifications to the tokenization process.

\begin{algorithm}[H]
\caption{PCA-Based Feature Creation for Time Series Signals}
\label{alg:feature_creation}
\begin{algorithmic}[1]
\State \textbf{Input:} Time series signals $S_{\text{RSSI}}$ and $S_{\text{SINR}}$
\State \textbf{Output:} Enhanced feature sets for RSSI and SINR
\For{each signal $S$ in $\{S_{\text{RSSI}}, S_{\text{SINR}}\}$}
 \State $X \gets \text{rolling\_window}(S, \text{size}=300)$ \Comment{Create samples, shape: $(|S|-299, 300)$}
 \State Define transformation set $\mathcal{T}$:
 \State \quad $T_1(X) = X$ \Comment{Original samples}
 \State \quad $T_2(X) = \text{moving\_average}(X, n=2)[:,2:]$ \Comment{Window 2, columns 2 to end}
 \State \quad $T_3(X) = \text{moving\_average}(X, n=3)[:,3:]$ \Comment{Window 3, columns 3 to end}
 \State \quad $T_4(X) = \text{moving\_average}(X, n=5)[:,5:]$ \Comment{Window 5, columns 5 to end}
 \State \quad $T_5(X) = X[:,::2]$ \Comment{Every 2nd point, start at 0}
 \State \quad $T_6(X) = X[:,1::2]$ \Comment{Every 2nd point, start at 1}
 \State \quad $T_7(X) = X[:,::3]$ \Comment{Every 3rd point, start at 0}
 \State \quad $T_8(X) = X[:,1::3]$ \Comment{Every 3rd point, start at 1}
 \State \quad $T_9(X) = X[:,2::3]$ \Comment{Every 3rd point, start at 2}
 \For{each transformation $T_i$ in $\mathcal{T}$}
 \State Fit $\text{PCA}_i$ on $T_i(X)$ with $n_{\text{components}}$ retaining 99\% variance
 \EndFor
 \State $X_{\text{pca}} \gets \text{concatenate}([\text{PCA}_i.\text{transform}(T_i(X))[:,:5] \text{ for } i = 1 \text{ to } 9], \text{axis}=1)$
 \State $\text{min}_X, \text{max}_X \gets \min(X), \max(X)$ \Comment{Global min and max of $X$}
 \State $X_{\text{pca}} \gets \text{scale}(X_{\text{pca}}, \text{range}=[\text{min}_X, \text{max}_X])$ \Comment{Scale each column}
 \State $X_{\text{enhanced}} \gets \text{concatenate}([X, X_{\text{pca}}], \text{axis}=1)$
\EndFor
\State \Return $X_{\text{enhanced}}$ for each signal
\end{algorithmic}
\end{algorithm}

\paragraph{\textbf{Tokenization}}

The data processing pipeline of the proposed transformer model begins with the collection of enhanced variants of RSSI and SINR derived via Algorithm \ref{alg:feature_creation}. Each enhanced signal is discretized into 50 equal-sized bins using percentile-based discretization, ensuring a uniform distribution across the data range. This process assigns a unique integer value to each bin, yielding 50 tokens per signal type. Consequently, two sets of 50 bins are produced—one for RSSI and one for SINR—resulting in a total of 100 distinct tokens representing the wireless signal characteristics.

To construct the token vocabulary, additional utility tokens are incorporated: \texttt{CLS} (assigned index 1), \texttt{MASK} (index 4), and binary classification tokens \texttt{NO\_ATTACK} (index 5) and \texttt{ATTACKED/JAMMED} (index 6). Other indices until 9 are reserved for potential future use, while the RSSI and SINR tokens span indices 9 to 109. This configuration results in a total vocabulary size of 110 tokens, with the maximum index capped at 110 for consistency across both LoS and NLoS scenarios. For training data, Time-Series Augmentation (TSA) techniques, inspired by \cite{Viana2024}, are applied to the discretized RSSI and SINR signals to enhance robustness. The input sequence is then formed by concatenating the \texttt{CLS} token, the discretized enhanced RSSI, the discretized enhanced SINR, and the label token. For LoS, this yields an input length of \(1 + 345 + 345 + 1 = 692\) tokens, while for NLoS, the length is \(1 + 345 + 309 + 1 = 656\) tokens, reflecting differences in signal characteristics between the two scenarios.

\begin{figure*}[!ht]
 \centering
 \includegraphics[width=1\textwidth]{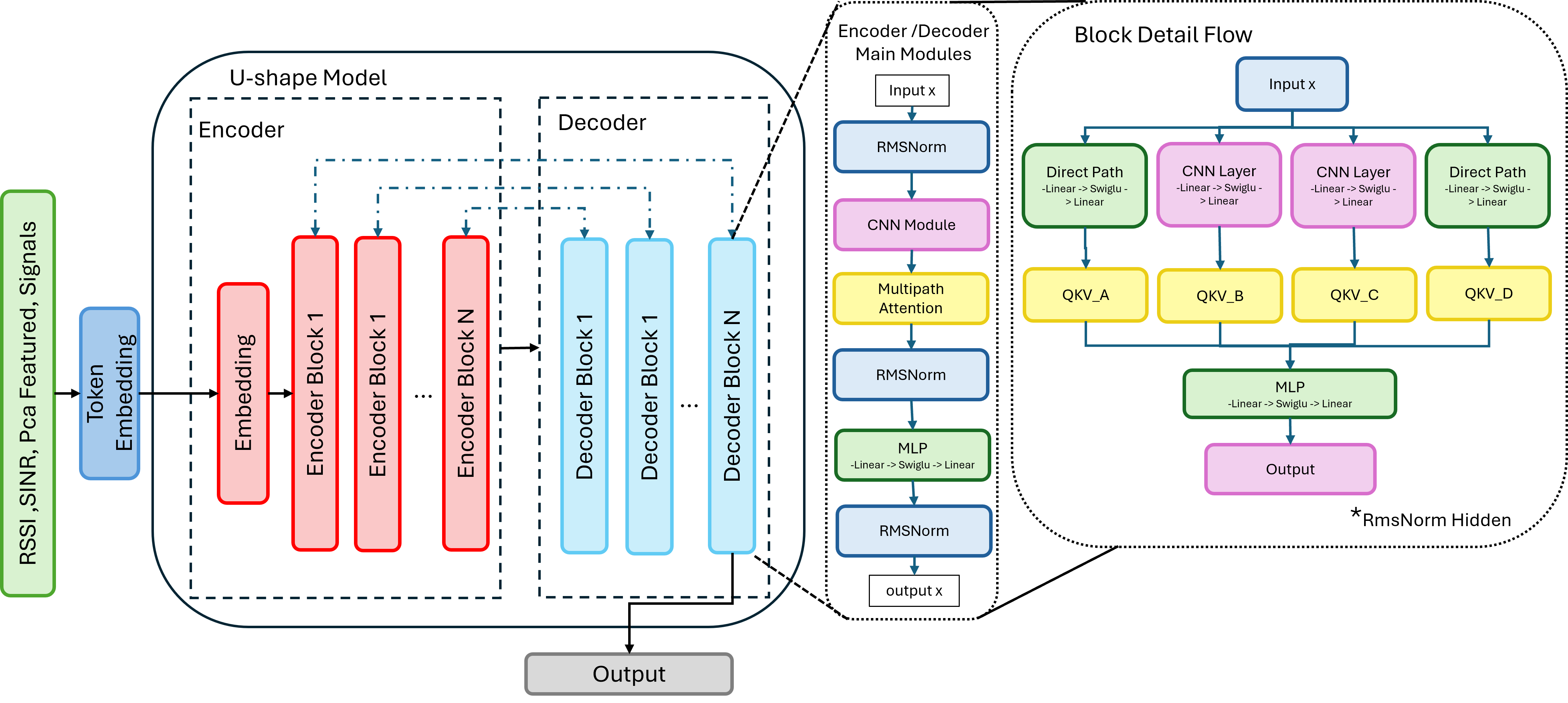}
 \caption{U\_Shaped Architecture}
 \label{fig:architecture}
\end{figure*}
 
\subsubsection{Deep Network Design}
Building upon prior work on multi-headed deep networks with attention mechanisms \cite{Viana2024}, we propose a U-shaped deep network architecture, illustrated in figure~\ref{fig:architecture}, designed to deliver state-of-the-art performance in NLoS scenarios without reliance on post-processing techniques. Inspired by the U-Net architecture \cite{Ronneberger2015}, this model incorporates modifications to integrate multi-headed attention and variable embedding dimensions across layers, enhancing its capability for robust signal analysis.

The architecture consists of an encoder pathway with three stages, a corresponding decoder pathway with three stages, and residual skip connections that link the two. These skip connections enable direct gradient propagation from deeper to shallower layers, mitigating gradient degradation during backpropagation. The model processes input features and generates output tokens that encapsulate predictions, achieving competitive performance without additional refinement steps.

Input processing begins with the integration of RSSI measurements, SINR values, and PCA-derived features. These inputs are transformed into high-dimensional representations via a token embedding layer, preparing them for subsequent transformer-based processing.

The encoder pathway, depicted in red in figure~\ref{fig:architecture}, comprises sequential blocks that progressively refine input representations. Each block integrates convolutional neural network (CNN) modules for local feature extraction, multi-headed attention mechanisms for selective signal focus, and root mean square normalization (RMSNorm) layers for training stability. These components are complemented by multi-layer perceptron (MLP) units featuring a Linear-SwiGLU-Linear activation sequence, forming a robust feature extraction pipeline.

The decoder pathway, shown in blue, mirrors the encoder's structure while incorporating skip connections from the corresponding encoder layers to preserve critical information. Symmetry between the encoder and decoder is maintained, with the decoder leveraging similar components to reconstruct and interpret encoded features for classification. This U-shaped configuration enhances multi-scale feature analysis within the transformer framework.

Within each block, parallel processing paths enhance feature extraction: direct Linear-SiLU-Linear transformations operate alongside CNN layers, while multiple query-key-value (QKV) attention mechanisms process distinct signal aspects concurrently. Inspired by the DIFF Transformer \cite{ye2025differential}, differential computations are applied to the attention outputs of specific heads (e.g., heads 6 and 7) to enrich feature representations. A final MLP layer integrates these diverse features for comprehensive analysis.

RMSNorm is employed at multiple stages to ensure consistent normalization and training stability, enabling the model to focus on pertinent signal characteristics while suppressing noise. Skip connections further preserve gradient flow, enhancing training efficiency. The architecture culminates in a classification layer that leverages these rich feature representations to accurately predict jamming threats, capturing both overt and subtle attack patterns.

This design excels in processing multiple input feature types simultaneously while preserving their interrelationships. The U-shaped structure facilitates hierarchical feature extraction, and skip connections ensure information retention across processing stages, making the model well-suited for complex signal analysis tasks.

\paragraph{Detailed Architecture Specification}
The encoder and decoder blocks operate at varying embedding dimensions, defined as follows:
\begin{itemize}
 \item \textbf{Encoder Dimensions:} 256, 128, 64
 \item \textbf{Decoder Dimensions:} 64, 128, 256
\end{itemize}
Each block comprises:
\begin{itemize}
 \item An RMSNorm layer for normalization.
 \item A CNN layer with kernel size 3 and padding 1 for local feature extraction.
 \item A Linear-SiLU-Linear transformation sequence.
 \item An 8-headed attention mechanism with differential computations.
 \item Residual connections paired with SwiGLU-activated MLP blocks.
\end{itemize}
The hierarchical reduction of embedding dimensions in the encoder is symmetrically reversed in the decoder, maintaining structural balance.

\paragraph{Attention Mechanism and Differentiation}
The multi-headed attention mechanism processes embeddings through distinct groups:
\begin{itemize}
 \item Heads 0 and 1 directly process normalized inputs.
 \item Heads 2 and 3 process CNN-transformed embeddings via an MLP.
 \item Heads 4 and 5 rely solely on CNN transformations.
 \item Heads 6 and 7 compute differences between normalized inputs and CNN-transformed outputs.
\end{itemize}
This configuration, combined with differential attention inspired by \cite{ye2025differential}, enhances the model's ability to capture nuanced signal variations, contributing to its effectiveness in identifying jamming patterns.

\subsubsection{\textbf{Proposed Training Algorithm}}
In this work, we propose a robust training framework for time series data that accelerates convergence and improves generalization. The framework integrates four key components: (i) a chunking strategy for data sampling, (ii) dynamic batch size scheduling combined with learning rate adjustment, (iii) an exponential moving average (EMA) for model weights with an integrated restoration mechanism, and (iv) mixed precision training with gradient clipping. These components are detailed in the following sections.

\paragraph{Chunking Strategy}
To mitigate temporal correlations and reduce training time, we adopt a chunking strategy that partitions the training dataset into a fixed number of chunks (e.g., 10). In each epoch, a subset is selected from the dataset by choosing every $n$th sample with an offset determined by a randomized permutation. This ensures that samples used in a mini-batch maintain a minimum temporal gap, thereby promoting diversity and reducing overfitting. The pseudocode for the chunking algorithm is provided in Algorithm~\ref{alg:chunking}.

\begin{algorithm}[H]
\caption{Chunking Strategy for Data Selection}
\label{alg:chunking}
\begin{algorithmic}[1]
\Require Number of chunks $n$, current epoch $e$, training dataset $D_{\text{train}}$
\State Set $n \gets 10$ \Comment{User-defined number of chunks}
\State Compute $s \gets e \mod n$
\If{$s = 0$}
 \State Generate a new random permutation $P$ of $\{0,1,\dots,n-1\}$
 \State \textbf{set} $s \gets 0$
\Else
 \State Set $s \gets P[s]$
\EndIf
\State Select subset: $D_{\text{subset}} \gets D_{\text{train}}[s\,:\,:\,n]$
\State \Return $D_{\text{subset}}$
\end{algorithmic}
\end{algorithm}

In each epoch, the subset $D_{\text{subset}}$ is used for training, ensuring that samples are minimally correlated (with a gap of at least 10 time steps) and that the overall data diversity is maintained.

\paragraph{Dynamic Batch Size Scheduling and Learning Rate Adjustment}
To optimize convergence, our framework employs a dynamic batch size scheduler that adjusts the effective batch size via gradient accumulation. Initially, the scheduler is deactivated. If validation loss and accuracy fail to improve over successive epochs, the scheduler increases the effective batch size by modifying the gradient accumulation steps. This approach enables the network to benefit from larger batch sizes without incurring additional memory overhead. Concurrently, a learning rate scheduler with a warmup phase (e.g., 8 epochs) adjusts the learning rate dynamically. The combination of these schedulers accelerates convergence and enhances validation performance.

\paragraph{Weight Moving Average and Restoration Mechanism}
To stabilize training and improve generalization, an exponential moving average (EMA) of the model parameters is maintained. When validation performance improves, the current model state is saved as the best checkpoint. The EMA is then updated according to
\[
W_{\text{new}} = \alpha \cdot W_{\text{prev}} + (1 - \alpha) \cdot W_{\text{current}},
\]
where $\alpha$ is a small factor (e.g., $\alpha=0.001$), $W_{\text{prev}}$ denotes the previously maintained weight vector, and $W_{\text{current}}$ represents the current weights. In cases where validation metrics degrade for two consecutive epochs (tracked via a restoration counter), the model is restored to the best checkpoint and a modified weight update is applied (with, for instance, $\alpha=0.005$) to prevent divergence. This restoration mechanism ensures that the training process remains stable despite fluctuations in performance.

\paragraph{Mixed Precision Training and Gradient Clipping}
To improve computational efficiency, the training process employs mixed precision training using PyTorch's automatic mixed precision (AMP) framework. Computations are performed in bfloat16 precision on CUDA devices, and a gradient scaler is used to avoid numerical underflow. Furthermore, gradient clipping with a maximum norm of 1.0 is applied to prevent exploding gradients, thereby ensuring stable and robust model updates.

\paragraph{Loss Function with Entropy Regularization}
Inspired by the regularized entropy objective in the Soft Actor-Critic (SAC) algorithm~\cite{sac2018}, the standard classification loss is augmented with an entropy regularizer. The modified loss function is defined as
\[
L = \frac{L_{\text{base}} - \lambda_{\text{entropy}}\, H(X)}{N_{\text{accum}}},
\]
where $L_{\text{base}}$ is the standard cross-entropy loss, 
\[
H(X) = -\sum_{i=1}^{n} P(x_i) \log P(x_i)
\]
denotes the entropy of the predicted probability distribution, $\lambda_{\text{entropy}}$ (e.g., 0.4) is a hyperparameter controlling the regularization strength, and $N_{\text{accum}}$ is the number of gradient accumulation steps. This entropy regularization encourages the model to avoid overconfident predictions, thereby promoting better generalization. In practice, if the computed loss is non-finite (i.e., NaN or infinity), the corresponding batch is skipped to ensure robustness.

\paragraph{Summary}
The integration of a data chunking strategy, dynamic batch size and learning rate scheduling, an EMA with a restoration mechanism, and mixed precision training with gradient clipping results in a comprehensive framework for training models on time series data. Experimental results demonstrate that this approach not only accelerates convergence but also enhances generalization, yielding improved validation and test performance.

\section{Results}

This section presents the results and insights gained from our UAV jamming detection algorithms experiment using \text{PCA-featured} (proposed), DNN, DNN+M1, DNN+M2, and XGB classifier algorithms from \cite{Viana2024}, CNN Architecture from \cite{9920202} along with details of the data used to train all the algorithms. Unless explicitly stated, the general network and U-shaped transformer model parameters employed in the experiment are described in Tables \ref{table:1} and \ref{table:2}, respectively.

\begin{table}[!ht]
\centering
\begin{threeparttable}

\begin{tabular}{@{}ll@{}}
\toprule
Scenario Parameters & \textbf{Values} \\ \midrule
Terrestrial Users & 0, 5, 10 \\
Authenticated UAVs & 1 \\
Small Cells & 10 \\
Small cell height & 10 m \\
Attackers & 0, 1, 2, 3, and 4 \\
Speeds & 10 m/s \\
Small cell power & 4 dBm \\
Authenticated UAV power & 2 dBm \\
Attackers power & 0, 2, 5, 10, and 20 dBm \\
Authenticated UAV position & URD* \\
Attackers position & URD* \\
Small cells position & URD* \\
Scenario & UMi \\
Distance & 100, 200, 500, and 1000 m\\
\bottomrule
\end{tabular}
 \begin{tablenotes}
 \small
 \item *URD - Uniformly Random Distributed.
 \end{tablenotes}
 \end{threeparttable}
\caption{Dataset Parameters. \cite{Viana2024}}
\label{table:1}
\end{table}
\vspace{-1mm}

\begin{table}[!ht]
\centering
\begin{tabular}{lc} 
\toprule
Parameter & \multicolumn{1}{c}{\textbf{LoS} and \textbf{NLoS}} \\ 
\midrule
Block size (LoS, NLoS) & \multicolumn{1}{c}{(692, 656)} \\
Layer number & \multicolumn{1}{c}{6} \\
Learning Rate & \multicolumn{1}{c}{1e-4} \\
Heads number & \multicolumn{1}{c}{8} \\
Vocab size & \multicolumn{1}{c}{110} \\
Encoder Embedding & \multicolumn{1}{c}{[256, 128, 64]} \\
Decoder Embedding & \multicolumn{1}{c}{[64, 128, 256]} \\
Dropout & \multicolumn{1}{c}{0.4} \\
Batch Size & \multicolumn{1}{c}{64} \\
Noise & \multicolumn{1}{c}{0.03} \\
(Rand. Mask Prob., Target Mask Prob.) (training) & \multicolumn{1}{c}{(0.25, 0.85)} \\
(Rand. Mask Prob., Target Mask Prob.) (prediction) & \multicolumn{1}{c}{(0.0, 1.0)} \\
Model Parameters & \multicolumn{1}{c}{2.2 M} \\
\bottomrule
\end{tabular}
\caption{ U-shaped Transformer Model configuration}
\label{table:2}
\end{table}
\vspace{-1mm}

\subsection{\textbf{Classification Comparison with Other Algorithms}}

\begin{figure*}[!ht]
 \centering
 \includegraphics[width=0.99\textwidth]{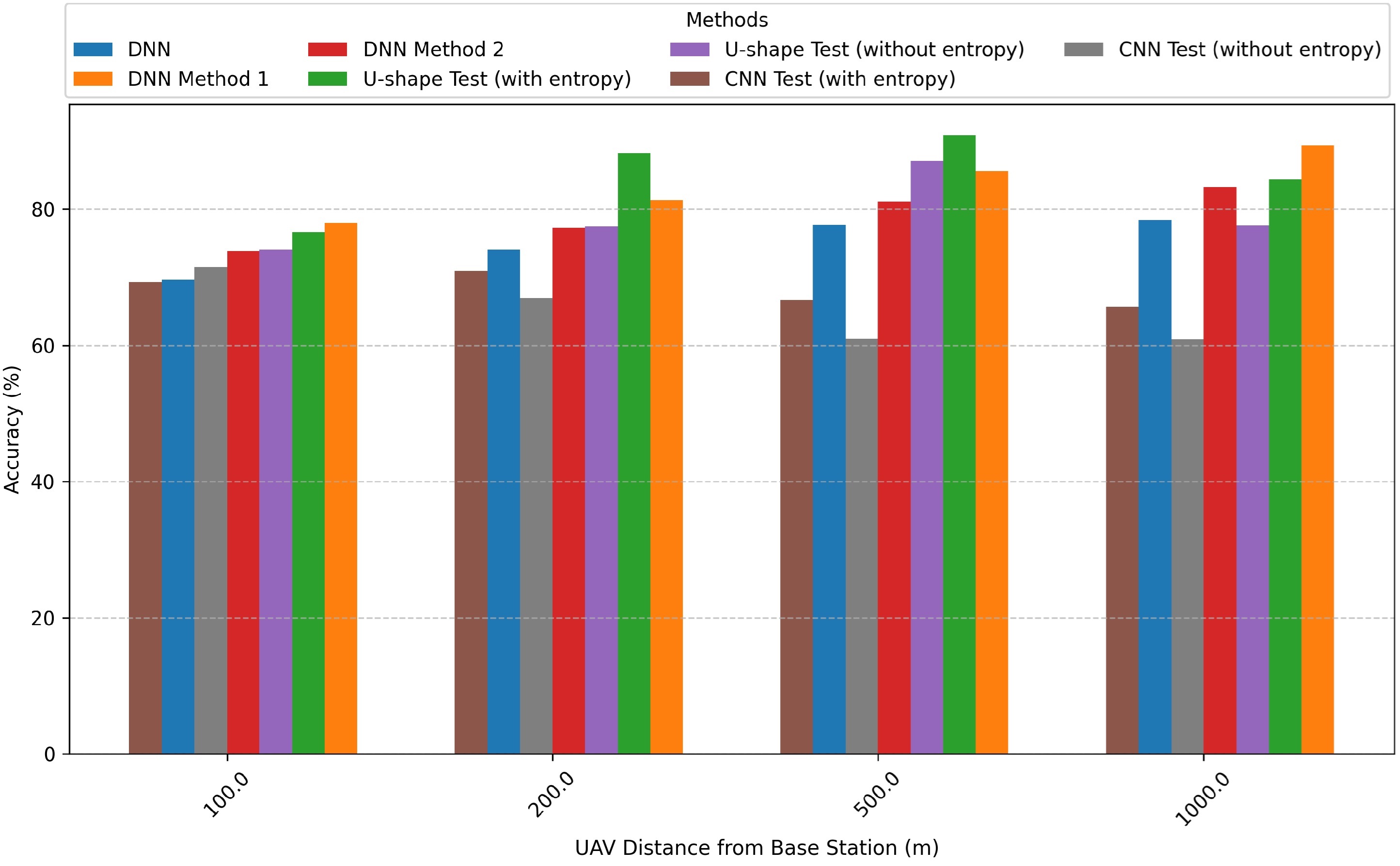}
 \caption{Comparison with Deep Learning Results Based UAV - BS Distance in NLoSConditions. The proposed model is accentuated in green for emphasis}
 \label{fig:DNNs_comparison}
\end{figure*}
\vspace{-1mm}

Table 3 presents the comparative performance of our proposed approaches against seven different baseline methods for classification accuracies in both NLoS and LoS scenarios. 

Our entropy-enhanced model ("Proposed + entropy") achieves the highest detection rate in NLoS conditions at 85.06\%, significantly outperforming all competing methods. This performance underscores the effectiveness of our uncertainty-based regularization technique in challenging signal environments where direct paths are obstructed. In LoS conditions, our entropy-enhanced model achieves a competitive 89.46\% detection rate, while the DNN+M2 approach from \cite{Viana2024} shows marginally better performance at 90.80\%.

The results demonstrate that entropy-based regularization consistently improves detection capabilities, as evidenced by the performance gap between our basic and entropy-enhanced models (79.10\% vs. 85.06\% in NLoS, and 87.37\% vs. 89.46\% in LoS). This pattern is also observed in CNN-based methods in \cite{9920202}, where entropy integration shows mixed effects depending on the signal environment. Notably, our proposed method without entropy regularization ("Proposed") reaches 87.37\% in LoS scenarios, outperforming both XGB (86.33\%) and CNN-based approaches (81.86\% and 79.89\%), further validating the effectiveness of our U-shaped transformer architecture even without the additional entropy component.
When comparing our approach with DNN variants in \cite{Viana2024}, we observe that our entropy-enhanced model provides more balanced performance across both LoS and NLoS conditions. While DNN+M1 and DNN+M2 achieve strong results in LoS scenarios (89.98\% and 90.80\% respectively), they demonstrate more significant performance degradation in the challenging NLoS environment (83.07\% and 79.00\%). This highlights the robustness of our proposed U-shaped transformer architecture with entropy regularization, which maintains high detection rates even under adverse signal conditions, making it a reliable solution for real-world wireless security applications such as the UAV non-line-sight applications.

\subsection{\textbf{Distance Based Accuracy Comparison}}

The bar chart in figure \ref{fig:DNNs_comparison} compares accuracy performance of seven neural network methods at four UAV distances from a base station: 100m, 200m, 500m, and 1000m. The methods include three DNN variants (standard DNN, DNN Method 1, DNN Method 2) and four testing approaches (U-shape and CNN tests, each with and without an entropy regularizer).

At 100m, all methods achieve between 65-80\% accuracy, with DNN Method 1 and U-shape Test with entropy showing slightly better performance. As distance increases to 200m, a performance gap emerges with U-shape Test with entropy reaching 88.2\%, while CNN Test without entropy drops to about 66.9\%.
The 500m distance marks the peak performance point for most methods, with U-shape Test with entropy exceeding 90.9\% accuracy, closely followed by U-shape Test without entropy at around 87.1\% and DNN Method 1 at approximately 85\%. Both DNN Method 2 and standard DNN also perform well at this distance, while CNN tests remain significantly lower.
At the maximum tested distance of 1000m, DNN Method 1 achieves the highest accuracy at approximately 89.3\%, followed by DNN Method 2 and U-shape Test with entropy at about 84\%. This suggests DNN Method 1 has superior performance at extreme distances, while CNN-based approaches consistently show the poorest results across all tested ranges.
The consistent superiority of entropy-incorporated methods across all distances indicates that entropy provides valuable information for the neural network classification process in this UAV application context.

\section{Conclusions}
This research introduces an transformer-based architecture for detecting jamming attacks in UAV-integrated 5G networks, enhanced by PCA features and an entropy-based regularizer. The inclusion of a cross-entropy loss function with an entropy regularizer effectively penalizes overconfident predictions, leading to significant improvements in detection performance. The model achieves the highest detection rate of 85.06\% in Non-Line-of-Sight (NLoS) scenarios, with entropy-enhanced models outperforming conventional machine learning methods, such as XGBoost, by approximately 4.5\%, and deep learning models by approximately 2\%. The model’s success can be attributed to its efficient processing of multidimensional signal features, such as RSSI and SINR, combined with self-attention mechanisms that capture complex temporal dependencies in wireless signal patterns.

The comparative performance of our approach highlights its robustness across both LoS and NLoS conditions. In NLoS scenarios, our entropy-enhanced model significantly outperforms all competing methods, while in LoS conditions, it remains highly competitive, demonstrating the effectiveness of our entropy regularization technique in challenging environments. These results underscore the potential of transformer-based architectures in advancing wireless security, particularly for UAV applications, where jamming threats are increasingly prevalent.

\begin{IEEEbiography}[{\includegraphics[width=1in, height=1.25in, clip, keepaspectratio]
{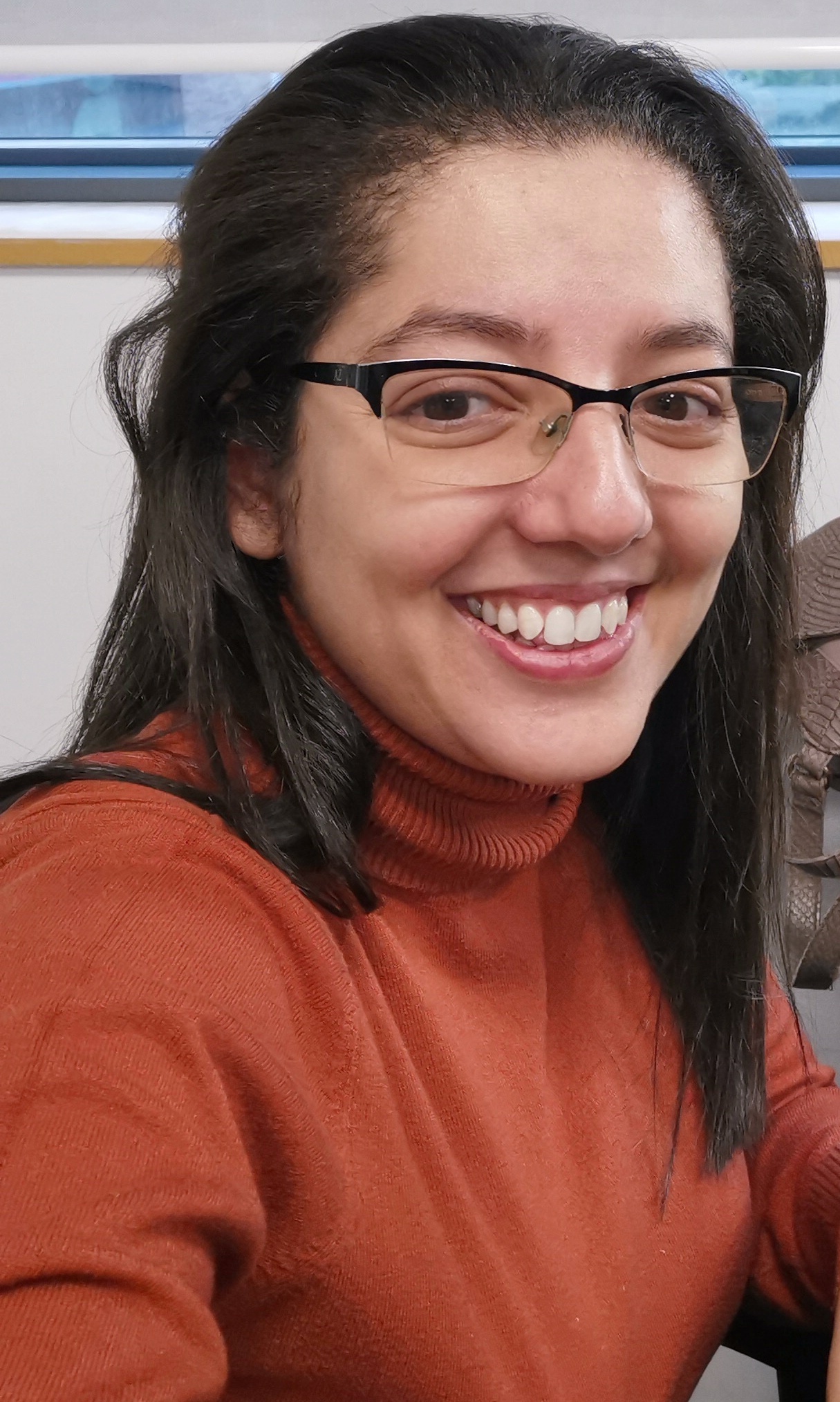}}]{Joseanne Viana} 
is a Ph.D. researcher at UC3M - Charles III University of Madrid. She received her bachelor’s degree in telecommunication engineering from the University of Campinas, Brazil. She is an Early-Stage Researcher in the project TeamUp5G, a European Training Network under the MSCA ITN of the European Commission’s Horizon 2020. Her research interests include wireless communications applied to interconnected systems such as UAVs, aerial vehicles, and non-terrestrial devices.
\end{IEEEbiography}

 \vskip -2\baselineskip plus -1fil
\begin{IEEEbiography}[{\includegraphics[width=1in, height=1.25in, clip, keepaspectratio]{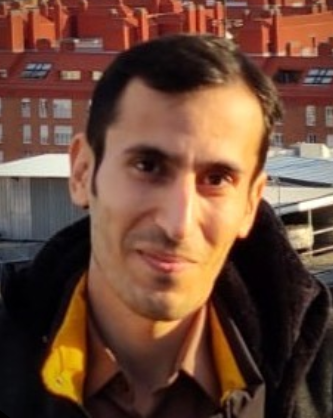}}]{Hamed Farkhari} 
is a Ph.D. researcher at ISCTE - Lisbon University Institute. He serves as an Early-Stage Researcher in the TeamUp5G group, a European Training Network under the Marie Skłodowska-Curie Actions (MSCA ITN) of the European Commission’s Horizon 2020 program. His research interests and work focus on cybersecurity, machine learning, deep learning, data science, meta-heuristic techniques, and optimization algorithms.
\end{IEEEbiography}
\vskip -2\baselineskip plus -1fil

\begin{IEEEbiography}[{\includegraphics[width=1in,height=1.25in,clip,keepaspectratio]{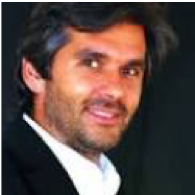}}]{Pedro Sebastião} received the Ph.D. degree in electrical and computer engineering from IST.He is currently a Professor with ISCTE-IUL’s Information Science and Technology Department, He is also the Board Director of AUDAX-ISCTE - Entrepreneurship and Innovation Center, ISCTE, responsible for the LABS LISBOA Incubator and Researcher at the Institute of Telecommunications. He has oriented several master’s dissertations and doctoral theses. It has several scientific, engineering and pedagogical awards. Also, he has organized or co-organized more than 55 national and international scientific conferences. He planned and developed several postgraduate courses in technologies and management, entrepreneurship and innovation and transfer of technology and innovation. He has supported several projects involving technology transfer and creation of start-ups and spinoffs of value to society and market. He developed his professional activity in the National Defense Industries, initially in the Office of Studies and later as the Board Director of the Quality Department of the Production of New Products and Technologies. He was also responsible for systems of communications technology in the Nokia-Siemens business area. His main researching interests are in monitoring, control and communications of drones, unmanned vehicles, planning tools, stochastic process (modeling and efficient simulations), the Internet of Things, and efficient communication systems.
\end{IEEEbiography}
\vskip -2\baselineskip plus -1fil

\begin{IEEEbiography}[{\includegraphics[width=1in,height=1.25in,clip,keepaspectratio]{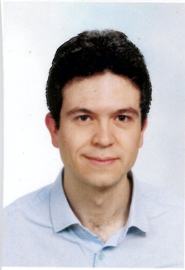}}]{Víctor P. Gil Jiménez} (Senior Member, IEEE) received a B.S. degree (Hons.) in Telecommunications from the University of Alcalá in 1998 and an M.S. degree (Hons.) in Telecommunications and a Ph.D. degree (Hons.) from Universidad Carlos III de Madrid in 2001 and 2005, respectively. In 1999, he was a Communications Staff member with the Spanish Antarctica Base. He visited the University of Leeds, U.K., in 2003, Chalmers Technical University, Sweden, in 2004, and the Instituto de Telecomunicações, Portugal, from 2008 to 2010. He is an Associate Professor with the Department of Signal Theory and Communications, Universidad Carlos III de Madrid. He has led several private and national Spanish projects and participated in various European and international projects. He holds one patent and has published over 80 journal articles/conference papers and nine book chapters. His research interests include advanced multicarrier systems for wireless radio, satellite, and visible light communications. He was the IEEE Spanish Communications and Signal Processing Joint Chapter Chair from 2015 to 2023. He received the Master Thesis Award and the Ph.D. Thesis Award from the Professional Association of Telecommunication Engineers of Spain in 1998 and 2006, respectively. 
\end{IEEEbiography}

\EOD

\end{document}